\documentclass[11pt,letterpaper]{article}
\usepackage[letterpaper]{geometry}
\usepackage{emnlp2018}
\aclfinalcopy
\usepackage{times}
\usepackage{latexsym}
\usepackage{amsmath}
\usepackage{appendix}
\usepackage{multirow}
\usepackage{url}

\usepackage{color}
\usepackage{comment}
\usepackage{graphicx}
\usepackage[T1]{fontenc}
\usepackage{subcaption}
\usepackage{enumitem}
\usepackage{algorithm}
\usepackage{booktabs}
\usepackage[noend]{algpseudocode}

\usepackage[all]{nowidow}

\newcommand{\example}[1]{\textit{#1}}
\newcommand{\gloss}[1]{``#1''}

\newcommand{\specialcell}[2][c]{%
  \begin{tabular}[#1]{@{}l@{}}#2\end{tabular}}

\newcommand{\var}[1]{\mathrm{#1}}
\newcommand{\set}[1]{\mathcal{#1}}
\newcommand{\model}[1]{\texttt{#1}}

\title{Making ``fetch'' happen: The influence of social and linguistic context on nonstandard word growth and decline}
\author{Ian Stewart and Jacob Eisenstein \\
School of Interactive Computing \\
Georgia Institute of Technology \\
Atlanta, GA 30318 \\
\url{{istewart6,jacobe}@gatech.edu}
}

\date{}

\begin{document}

\maketitle

\begin{abstract}
In an online community, new words come and go: today's \example{haha} may be replaced by tomorrow's \example{lol}. 
Changes in online writing are usually studied as a social process, with innovations diffusing through a network of individuals in a speech community.
But unlike other types of innovation, language change is shaped and constrained by the grammatical system in which it takes part.
To investigate the role of social and structural factors in language change, we undertake a large-scale analysis of the frequencies of nonstandard words in Reddit.
Dissemination across many linguistic contexts is a predictor of success: words that appear in more linguistic contexts grow faster and survive longer.
Furthermore, social dissemination plays a less important role in explaining word growth and decline than previously hypothesized.

\end{abstract}

\section{Introduction}

\textit{Stop trying to make ``fetch'' happen! It's not going to happen!} -- Regina George (\emph{Mean Girls}, 2005) \vspace{8pt}

With the fast-paced and ephemeral nature of online discourse, language change in online writing is both prevalent~\cite{androutsopoulos2011} and noticeable~\cite{squires2010enregistering}. 
In social media, new words emerge constantly to replace even basic expressions such as laughter: today's \example{haha} is tomorrow's \example{lol}~\cite{tagliamonte2008}. 
Why do some nonstandard words, like \example{lol}, succeed and spread to new contexts, while others, like \example{fetch}, fail to catch on? 
Can a word's growth be predicted from patterns of usage during its early days?

Language change
can be treated like other social innovations, such as the spread of hyperlinks~\cite{bakshy2011everyone} or hashtags~\cite{romero2011differences,tsur2015}.
A key aspect of the adoption of a new practice is its \emph{dissemination}: is it used by many people, and in many social contexts?
High dissemination enables words to achieve greater exposure among social groups~\cite{altmann2011}, and may signal that the innovation is positively evaluated.

In addition to social constraints, language change is also shaped by grammatical constraints~\cite{darcy2015}.
New words and phrases rarely change the rules of the game but must instead find their place in a competitive ecosystem with finely-differentiated linguistic roles, or ``niches''~\cite{macwhinney1989}.
Some words become valid in a broad range of linguistic contexts, while others remain bound to a small number of fixed expressions.
We therefore posit a structural analogue to social dissemination, which we call \emph{linguistic dissemination}.

We compare the fates of such words to determine how linguistic and social dissemination each relate to word growth, focusing on the adoption of nonstandard words in the popular online community Reddit.
The following hypotheses are evaluated:
\vspace{-1pt}
\begin{itemize}
  \setlength\itemsep{0pt}
  \setlength\parskip{0pt}
\item \textbf{H1: Nonstandard words with higher initial social dissemination are more likely to grow.} 
Following the intuition that words require a large social base to succeed, we hypothesize a positive correlation between social dissemination and word growth. 
\item \textbf{H2-weak: Nonstandard words with higher linguistic dissemination in the early phase of their history are more likely to grow.} 
This follows from work in corpus linguistics showing that words and grammatical patterns with a higher diversity of collocations are more likely to be adopted~\cite{ito2003,partington1993}.
\item \textbf{H2-strong: Nonstandard words with higher linguistic dissemination are more likely to grow, even after controlling for social dissemination.}
This follows from the intuition that linguistic context and social context contribute differently to word growth. 
\end{itemize}

To address H2, we develop a novel metric for characterizing linguistic dissemination, by comparing the observed number of $n$-gram contexts to the number of contexts that would be predicted based on frequency alone.
Our analysis of word growth and decline includes: (1) prediction of frequency change in growth words (as in prior work); (2) causal inference of the influence of dissemination on probability of word growth; (3) binary prediction of future growth versus decline; and (4) survival analysis, to determine the factors that predict when a word's popularity begins to decline.
All tests indicate that linguistic dissemination plays an important role in explaining the growth and decline of nonstandard words.

%

\section{Related Work}

\paragraph{Lexical change online}

Language changes constantly, and one of the most notable forms of change is the adoption of new words~\cite{metcalf2004}, sometimes referred to as lexical entrenchment~\cite{chesley2010}.
New nonstandard words may arise through the mutation of existing forms by processes such as truncation (e.g, \example{favorite} to \example{fave};\nocite{grieve2016} Grieve et al., 2016) and blending (e.g., \example{web}+\example{log} to \example{weblog} to \example{blog};\nocite{cook2010blend} Cook and Stevenson, 2010). 
The fast pace and interconnected nature of online communication is particularly conducive to innovation, and social media provides a ``birds-eye view'' on the process of change~\cite{danescu2013,kershaw2016,tsur2015}.

The most closely related work is a contemporaneous study that explored the role of weak social ties in the dissemination of linguistic innovations on Reddit, which also proposed the task of quantitatively predicting the success or failure of lexical innovations~\cite{tredici2018}.
One distinguishing feature of our work is the emphasis on \emph{linguistic} (rather than social) context in explaining these successes and failures. 
In addition to predicting the binary distinction between success and failure, we also take on the more fine-grained task of predicting the length of time that each nonstandard word will survive.

\paragraph{Social dissemination}
Language changes as a result of transmission across generations~\cite{labov2007} as well as diffusion across individuals and social groups~\cite{bucholtz1999}.
Such diffusion can be quantified with \emph{social dissemination}, which \newcite{altmann2011} define as the count of social units (e.g., users) who have adopted a word, normalized by the expected count under a null model in which the word is used with equal frequency across the entire population.
\newcite{altmann2011} use dissemination of words across forum users and threads to predict the words' change in frequency in Usenet, finding a positive correlation between frequency change and both kinds of social dissemination.
In contrast, \newcite{garley2012} use the same metric to predict the growth of English loanwords on German hip-hop forums, and find that social dissemination has less predictive power than expected.
We seek to replicate these prior findings, and to extend them to the broader context of Reddit.

\paragraph{Linguistic dissemination}
In historical linguistics, the distribution of a new word or construction across lexical contexts can signal future growth~\cite{partington1993}.
Furthermore, grammatical and lexical factors can explain a speaker's choice of linguistic variant~\cite{ito2003,cacoullos2009} and can provide more insight than social factors alone.
Our study proposes a generalizable method of measuring the dissemination of a word across lexical contexts with \emph{linguistic} dissemination and compares social and linguistic dissemination as predictors of language change.
\section{Data}

Our study examines the adoption of words on social media, and we focus on Reddit as a source of language change. 
Reddit is a social content sharing site separated into distinct sub-communities or ``subreddits'' that center around particular topics~\cite{gilbert2013}. 
Reddit is a socially diverse and dynamic online platform, making it an ideal environment for research on language change~\cite{kershaw2016}.
Furthermore, because Reddit data is publicly available we expect that this study can be more readily replicated than a similar study on other platforms such as Facebook or Twitter, whose data is less easily obtained.

We analyze a set of public monthly Reddit comments\footnote{From \url{http://files.pushshift.io/reddit/comments/} (Accessed 1 October 2016).} posted between 1 June 2013 and 31 May 2016, totalling $T=36$ months of data.
This dataset has been analyzed in prior work~\cite{hessel2016,tan2015} and has been noted to have some missing data~\cite{gaffney2018}, although this issue should not affect our analysis.
To reduce noise in the data, we filter all comments generated by known bots and spam users\footnote{The same list used in~\newcite{tan2015}: \url{https://chenhaot.com/data/multi-community/README.txt} (Accessed 1 October 2016).} 
and filter all comments created in well-known non-English subreddits.\footnote{We randomly sampled 100 posts from the top 500 subreddits and labelled a subreddit as non-English if fewer than 90\% of its posts were identified by \texttt{langid.py}~\cite{lui2012} as English.}
The final data collected is summarized in Table \ref{tab:data_summary}.

\begin{table}
\small
\centering
  \begin{tabular}{l r r}
    \toprule
    & Total & Monthly mean \\
    \midrule
    Comments   & 1,625,271,269 & 45,146,424 \\
    Tokens     & 56,674,728,199 & 1,574,298,006 \\
    Subreddits & 333,874 & 48,786 \\
    Users      & 14,556,010     & 2,302,812 \\
    Threads      & 102,908,726     & 3,079,780 \\
    \bottomrule
    \end{tabular}
    \caption{Data summary statistics.}
    \label{tab:data_summary}
\end{table}

%

We replace all references to subreddits and users (marked by the convention \example{r/subreddit} and \example{u/user}) with \example{r/SUB} and \example{u/USER} tokens, and all hyperlinks with a \example{URL} token. 
We also reduce all repeated character sequences to a maximum length of three (e.g., \example{loooool} to \example{loool}).
The final vocabulary includes the top 100,000 words by frequency.\footnote{We restricted the vocabulary because of the qualitative analysis required to identify nonstandard words.}
We replace all OOV words with UNK tokens, which comprise 3.95\% of the total tokens.

\subsection{Finding growth words}
\label{subsec:growth_words}


Our work seeks to study the growth of nonstandard words, which we identify manually instead of relying on pre-determined lists~\cite{tredici2018}.
%
To detect such words, we first compute the Spearman correlation coefficient 
between the time steps $\{1...T\}$ and each word $w$'s frequency time series $f^{(w)}_{(1:T)}$ (frequency normalized and log-transformed).
The Spearman correlation coefficient captures monotonic, gradual growth that characterizes the adoption of nonstandard words~\cite{grieve2016,kershaw2016}.

The first set of words is filtered by a Spearman correlation coefficient above the $85^{\text{th}}$ percentile ($N=15,017$).
From this set of words, one of the authors manually identified 1,120 words in set $\set{G}$ (``growth'') that are neither proper nouns (\example{berniebot, killary, drumpf}) nor standard words (\example{election, voting}).\footnote{Code and word lists available at: \\ \url{https://github.com/ianbstewart/nonstandard_word_dissemination}.} 
These words were removed because their growth may be due to exogenous influence.
A ``standard'' word is one that can plausibly be found in a newspaper article, which follows from the common understanding of newspaper text as a more formal and standard register.
Therefore, a ``nonstandard'' word is one that cannot plausibly be found in a newspaper article, a judgment often used by linguists to determine what counts as slang~\cite{dumas1978}.
In ambiguous cases, one of the authors inspected a sample of comments that included the word.
We validate this process by having both authors annotate the top 200 growth candidates for standard/proper versus nonstandard (binary), obtaining inter-annotator agreement of $\kappa$=0.79.

\subsection{Finding decline words}
To determine what makes the growth words successful, we need a control group of ``decline'' words, which are briefly adopted and later abandoned.
Although these words may have been successful before the time period investigated, their decline phase makes them a useful comparison for the growth words. 
We find such words by fitting two parametric models to the frequency series.

\paragraph{Piecewise linear fit}
We fit a two-phase piecewise linear regression on each word's frequency time series $f_{(1:T)}$, which splits the time series into $f_{(1:\hat{t})}$ and $f_{(\hat{t}+1:T)}$.
The goal is to select a split point $\hat{t}$ to minimize the sum of the squared error between observed frequency $f$ and predicted frequency $\hat{f}$:
\begin{equation}
\hat{f}(m_{1}, m_{2}, b,  t) = 
\begin{cases}
b + m_{1}t & t \leq \hat{t} \\
b + m_1 \hat{t} + m_{2}(t - \hat{t}) & t > \hat{t},
\end{cases}
\end{equation}
where $b$ is the intercept, $m_{1}$ is the slope of the first phase, and $m_{2}$ is the slope of the second phase.
Decline words $\set{D}_{p}$ (``piecewise decline'') display growth in the first phase ($m_{1} > 0$), decline in the second phase ($m_{2} < 0$), and a strong fit between observed and predicted data, indicated by $R^{2}(f, \hat{f})$ above the $85^{th}$ percentile (36.1\%); this filtering yields 14,995 candidates.

\paragraph{Logistic fit}
To account for smoother growth-decline trajectories, we also fit the growth curve to a logistic distribution, 
which is a continuous unimodal distribution with support over the non-negative reals. 
We identify the set of candidates $\set{D}_{l}$ (``logistic decline'') as words with a strong fit to this distribution, as indicated by $R^{2}$ above the $99^{th}$ percentile (82.4\%), yielding 998 candidates. 
The logistic word set partially overlaps with the piecewise set, because some words' frequency time series show a strong fit to both the piecewise function and the logistic distribution.


\paragraph{Combined set} 
We combine the sets $\mathcal{D}_{p}$ and $\mathcal{D}_{l}$ to produce a set of decline word candidates ($N=15,665$).
Next, we filter this combined set to exclude standard words and proper nouns, yielding a total of 530 decline words in set $\mathcal{D}$.
Each word is assigned a split point $\hat{t}$ based on the estimated time of switch between the growth phase and the decline phase, which is the split point $\hat{t}$ for piecewise decline words and the center of the logistic distribution $\hat{\mu}$ for the logistic decline words.

\begin{table}
\small
\centering
\begin{tabular}{l p{4.7cm}}
  \toprule
  Word set & Examples \\
  \midrule
$\mathcal{G}$ & \example{idk, lmao, shitpost, tbh, tho} \\
$\mathcal{D}_{l}$ & \example{atty, eyebleach, iifym, obeasts, trashy} \\
$\mathcal{D}_{p}$ & \example{brojob, nparent, rekd, terpers, wot} \\
\bottomrule
\end{tabular}
\caption{Examples of nonstandard words in all word sets: growth ($\mathcal{G}$), logistic decline ($\mathcal{D}_{l}$) and piecewise decline ($\mathcal{D}_{p}$).}
\label{tab:example_growth_decline_words}
\end{table}

Examples of both growth and decline words are shown in \autoref{tab:example_growth_decline_words}. 
The growth words include several acronyms (\example{tbh}, \gloss{to be honest}; \example{lmao}, \gloss{laughing my ass off}), while the decline words include clippings (\example{atty}, \gloss{atomizer}), respellings (\example{rekd}, \gloss{wrecked}; \example{wot}, \gloss{what}) and compounds (\example{nparent}, \gloss{narcissistic parent}).

\begin{table}
\small
\centering
\begin{tabular}{l p{1.1cm} p{1.1cm} p{1.1cm} p{0.9cm} l}
\toprule
 & Clipping & Compound & Respelling & Other & Total \\
\midrule
$\mathcal{G}$ & 198 (17.7\%) & 334 (29.8\%) & 83 (7.4\%) & 505 (45.1\%) & 1,120 \\ 
  $\mathcal{D}$ & 53 (10.0\%) & 100 (18.9\%) & 108 (20.4\%) & 269 (50.8\%) & 530 \\
  \bottomrule
\end{tabular}
\caption{Word formation category counts in growth ($\mathcal{G}$) and decline ($\mathcal{D}$) word sets.}
\label{fig:word_formation_category_counts}
\end{table}

We also provide a distribution of the words across word generation categories in \autoref{fig:word_formation_category_counts}, including compounds and clippings in similar proportions to prior work~\cite{kulkarni2018}.
Because the growth and decline words exhibit similar proportions of category counts, we do not expect that this will be a significant confound in differentiating growth from decline.




\section{Predictors}

We now outline the predictors used to measure the degree of \textbf{social} and \textbf{linguistic} dissemination in the growth and decline words.

\subsection{Social dissemination}

We rely on the dissemination metric proposed by~\newcite{altmann2011} to measure the degree to which a word occupies a specific social niche (e.g., low dissemination implies limited niche).
To compute user dissemination $D^U$ for word $w$ at time $t$, we first compute the number of individual users who used word $w$ at time $t$, written $U^{(w)}_t$. 
We then compare this with the expectation $\tilde{U}^{(w)}_t$ under a model in which word frequency is identical across all users.
The user dissemination is the log ratio,
\begin{equation}
\log \frac{U_{t}^{(w)}}{\tilde{U}_{t}^{(w)}} = \log U^{(w)}_{t} - \log \tilde{U}^{(w)}_{t}.
\end{equation}

Following \newcite{altmann2011}, the expected count $\tilde{U}_t^{(w)}$ is computed as,
\begin{equation}
\tilde{U}_{t}^{(w)} = \sum_{u \in \mathcal{U}_{t}}(1 - e^{-f_{t}^{(w)}m_{t}^{(u)}}),
\end{equation}
where $m_{t}^{(u)}$ equals the total number of words contributed by user $u$ in month $t$, and $\mathcal{U}_{t}$ is the set of all users active in month $t$. 
This corresponds to a model in which each token from a user has identical likelihood $f_t^{(w)}$ of being word $w$.
In this way, we compute dissemination for all users ($D^{U}$), subreddits ($D^{S}$) and threads ($D^{T}$) for each month $t \in \{1 ... T\}$.

\subsection{Linguistic dissemination}
\label{sec:linguistic_dissemination}

Linguistic dissemination captures the diversity of linguistic contexts in which a word appears, as measured by unique $n$-gram counts.
We compute the log count of unique trigram\footnote{Pilot analysis with bigram contexts gave similar results.} contexts for all words ($C^{3}$) using all possible trigram positions: in the sentence ``\example{that's cool af haha}'', the term \example{af} appears in three unique trigrams, \example{that's cool af, cool af haha, af haha <END>}.

The unique log number of trigram contexts is strongly correlated with log word frequency ($\rho(C^{3}, f) = 0.904$), as implied by Heaps' law~\cite{egghe2007}.
We therefore adjust this statistic by comparing with its expected value $\tilde{C}^{3}$.
At each timestep $t$, we fit a linear regression between log-frequency and log-unique $n$-gram counts, and then compute the residual between the observed log count of unique trigrams and its expectation, $D^{L} = C^{3}_t - \tilde{C}^{3}_t$.
The residual $D^{L}$, or \emph{linguistic dissemination}, identifies words with a higher or lower number of lexical contexts than expected.

\begin{figure}[t!]
\centering
\includegraphics[width=\columnwidth]{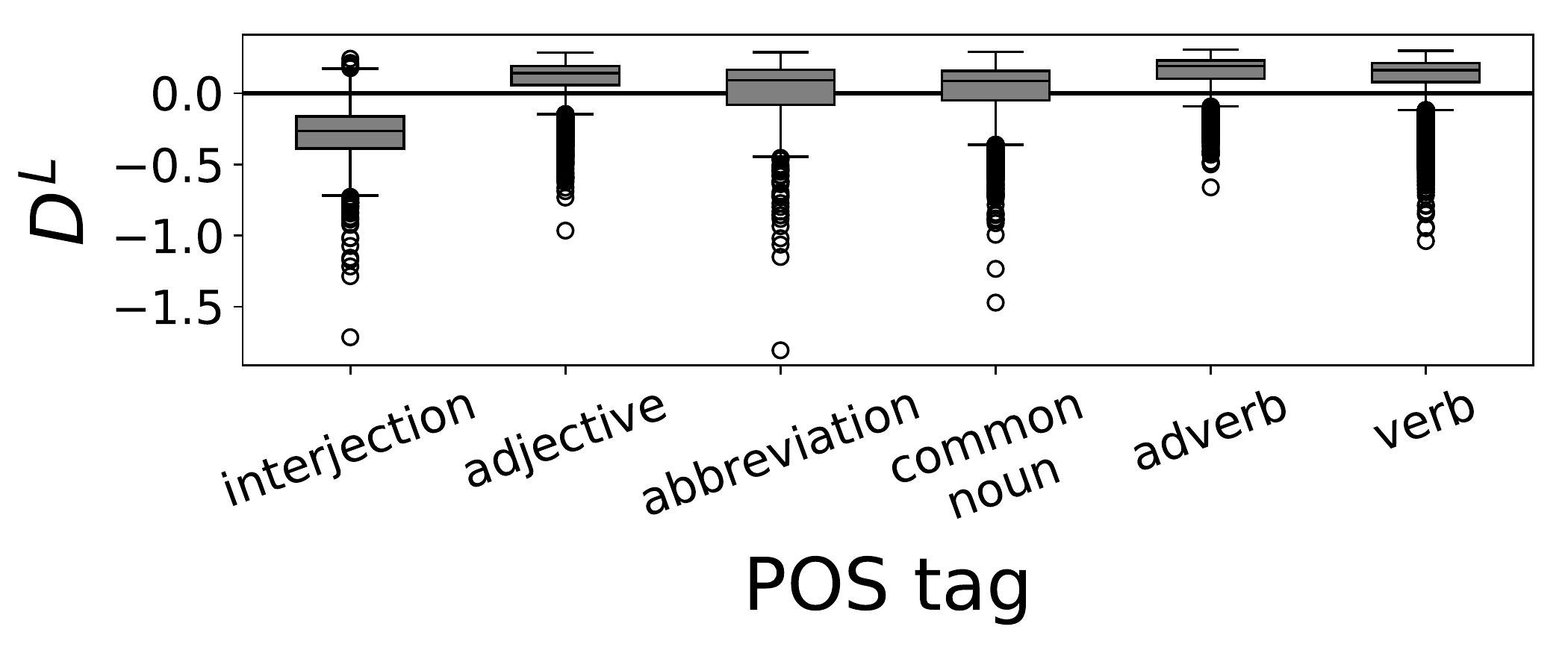}
\caption{Distribution of mean linguistic dissemination ($D^{L}$) across part of speech groups.}
\label{fig:pos-cd-dist}
\end{figure}

Linguistic dissemination can separate words by grammatical category, as shown in \autoref{fig:pos-cd-dist} where the mean $D^{L}$ values are computed for words across common part-of-speech categories.
Part-of-speech tags were computed over the entire corpus using a Twitter-based tagger~\cite{gimpel2011}, and each word type was assigned the most likely POS tag to provide an approximate distribution of tags over the vocabulary.
Interjections have a lower median $D^{L}$ than other word categories due to the tendency of interjections to occur in limited lexical contexts.
Conversely, verbs have  a higher median $D^{L}$ due to the flexibility of verbs' arguments (e.g., subject and object may both be open-class nouns).

\section{Results}
The hypotheses about social and linguistic dissemination are tested under four analyses: correlation against frequency change in growth words; causal inference on probability of word growth; binary prediction of word growth; and survival analysis of decline words.

\subsection{Correlational analysis}
\label{subsec:relative-importance}
To test the relative importance of the linguistic and social context on word growth, we correlate these metrics with frequency change ($\Delta_{f_{t}} = f_{t} - f_{t-k}$) across all growth words. 
This replicates the methodology in prior work by \newcite{altmann2011} and \newcite{garley2012}, who analyzed different internet forums.
Focusing on long-term change with $k=12$ (one year) and $k=24$ (two years), 
we compute the proportion of variance in frequency change explained by the covariates using a relative importance regression~\cite{kruskal1987}.\footnote{Relative importance regression implemented in the \emph{relaimpo} package in R: \url{https://cran.r-project.org/package=relaimpo}}

\begin{table}[t!]
\small
\centering
\begin{tabular}{l r r }
  \toprule
  ~ & \specialcell{Variance \\ explained} & Lower, upper 95\% \\
  \midrule
$f_{t-12}$ & 10.8\% & [10.2\%, 11.5\%]  \\ 
$D^{L}_{t-12}$ & 0.584\% & [0.461\%, 0.777\%]  \\ 
$D^{U}_{t-12}$ & 0.307\% & [0.251\%, 0.398\%] \\ 
$D^{S}_{t-12}$ & 0.120\% & [0.0852\%, 0.191\%] \\ 
$D^{T}_{t-12}$ & 0.246\% & [0.171\%, 0.379\%]  \\[2ex]
$f_{t-24}$ & 21.4\% & [20.4\%, 22.4\%] \\ 
$D^{L}_{t-24}$ & 1.29\% & [1.05\%, 1.64\%] \\ 
$D^{U}_{t-24}$ & 0.400\% & [0.346\%, 0.493\%] \\ 
$D^{S}_{t-24}$ & 0.287\%  & [0.201\%, 0.392\%] \\ 
$D^{T}_{t-24}$ & 0.272\% & [0.226\%, 0.380\%] \\ \bottomrule
\end{tabular}

\caption{Percent of variance explained in frequency change, computed over all growth words $\set{G}$.
$N=26,880$ for $k=12$, $N=13,440$ for $k=24$.} 
\label{tab:relative-importance-test}
\end{table}

The results of the regression are shown in \autoref{tab:relative-importance-test}. 
All predictors have relative importance greater than zero, according to a bootstrap method to produce confidence intervals~\cite{tonidandel2009}. 
Frequency is the strongest predictor ($f_{t-12}, f_{t-24}$), because words with low initial frequency often show the most frequency change.
In both short- and long-term prediction, linguistic dissemination ($D^{L}_{t-12}, D^{L}_{t-24}$) has a higher relative importance than each of the social dissemination metrics.
The social dissemination metrics have less explanatory power, in comparison with the other predictors and in comparison to the prior results of \newcite{garley2012}, who found $1.5\%$ of variance explained by $D^{U}$ and $1.9\%$ for $D^{T}$ at $k=24$.
Our results were robust to the exclusion of the predictor $D^L$, meaning that a model with only the social dissemination metrics as predictors resulted in a similar proportion of variance explained.
The weakness of social dissemination could be due to the fragmented nature of Reddit, compared to more intra-connected forums. 
Since users and threads are spread across many different subreddits, and users may not visit multiple subreddits, a higher social dissemination for a particular word may not lead to immediate growth.

\subsection{Causal analysis}
\label{sec:results-causal}
While correlation can help explain the relationship between dissemination and frequency change, it only addresses the weak version of H2: it does not distinguish the causal impact of linguistic and social dissemination. 
To test the strong version of H2, we turn to a causal analysis, in which the \emph{outcome} is whether a nonstandard word grows or declines, the \emph{treatment} is a single dissemination metric such as linguistic dissemination, and the \emph{covariates} are the remaining dissemination metrics. 
The goal of this analysis is to test the impact of each dissemination metric, while holding the others constant. 

\begin{figure*}
\includegraphics[width=\textwidth]{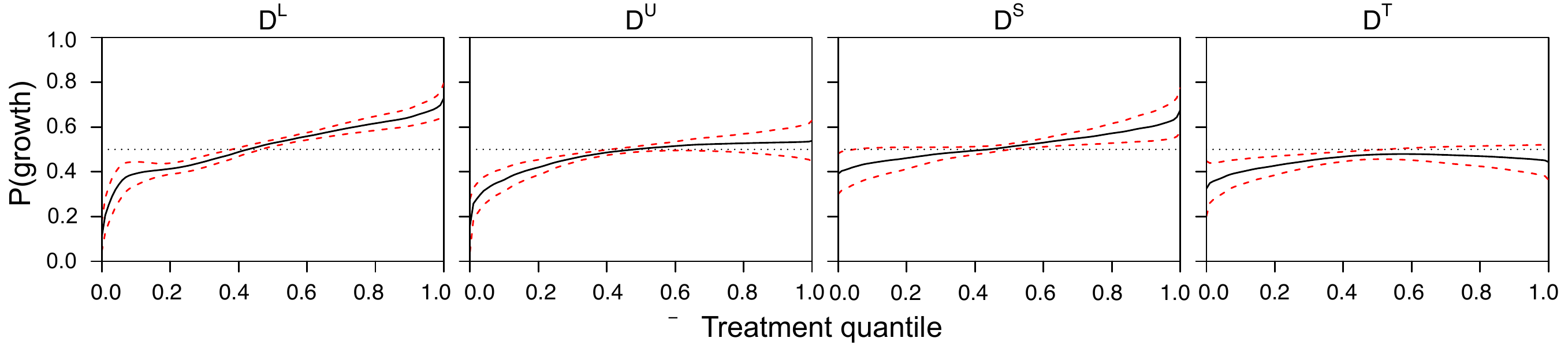}
\caption{Average dose response function for all treatment variables, where outcome is probability of word growth. 95\% confidence intervals plotted in red, chance rate of 50\% marked with dotted black line.}
\label{fig:ADRF_curves}
\end{figure*}

Causal inference typically uses a binary treatment/control distinction~\cite{angrist1996}, but in this case the treatment is continuous. 
We therefore turn to an adapted model known as the \emph{average dose response function} to measure the causal impact of dissemination~\cite{imbens2000}.
To explain the procedure for estimating the average dose response, we adopt the following terminology: $Z$ for treatment variable, $X$ for covariates, $Y$ for outcome.\footnote{Average dose response function implemented in the \emph{causaldrf} package in R: \url{https://cran.r-project.org/package=causaldrf}}
\begin{enumerate}
\setlength{\itemsep}{1pt}
\item A linear model is fit to estimate the treatment from the covariates,
  \begin{equation}
    Z_{i} \mid X_{i} \sim \mathcal{N}(\beta^{\top} X_{i},\sigma^{2}).
  \end{equation}
  The output of this estimation procedure is a vector of weights $\hat{\beta}$ and a variance $\hat{\sigma}^2$. 
\item The generalized propensity score (GPS) $R$ is the likelihood of observing the treatment given the covariates, $P(Z_{i} \mid X_{i})$. It is computed from the parameters estimated in the previous step:
  \begin{equation}
    \hat{R}_i = \frac{1}{\sqrt{2\pi\hat{\sigma}^{2}}} \exp \left(-\frac{(Z_{i} - \hat{\beta}^{\top}X_{i})^{2}}{2\hat{\sigma}^{2}}\right).
  \end{equation}
\item A logistic model is fit to predict the outcome $Y_{i}$ using the treatment $Z_{i}$ and the GPS $\hat{R}_i$:
  \begin{equation}
    \hat{Y}_{i} = \text{Logistic}(\hat{\alpha}_{0} + \hat{\alpha}_{1}Z_{i} + \hat{\alpha}_{2}\hat{R}_i).
  \end{equation}
  This involves estimating the parameters $\{\hat{\alpha}_0, \hat{\alpha}_1, \hat{\alpha}_2.\}$
By incorporating the generalized propensity score $\hat{R}_i$ into this predictive model over the outcome, it is possible to isolate the causal effect of the treatment from the other covariates~\citep{hirano2004}.
\item The range of treatments is divided into levels (quantiles). The \emph{average dose response} for a given treatment level $s_{z}$ is the mean estimated outcome for all instances at that treatment level,
\begin{equation}
\hat{\mu}(s_{z}) = \frac{1}{| s_{z} |} \sum_{z_{i} \in s_{z}} \hat{Y}_{i}.
\end{equation}
The average dose response function is then plotted for all treatment levels.
\end{enumerate}

Each dissemination metric is considered separately as a treatment. 
We consider all other dissemination metrics and frequency as covariates: e.g., for treatment variable $D^{L}$, the covariates are set to $[f,D^{U},D^{S},D^{T}]$.
We bootstrap the above process 100 times with different samples to produce confidence intervals.
To balance the outcome classes, we sample an equal number of growth and decline words for each bootstrap iteration.

The average dose response function curves in \autoref{fig:ADRF_curves} show that linguistic dissemination ($D^{L}$) produces the most dramatic increase in word growth probability.
For linguistic dissemination, the lowest treatment quantile (0\%-10\%) yields a growth probability below 40\% (significantly less than chance), as compared to the highest treatment quantile (90-100\%), which yields a growth probability nearly at 70\% (significantly greater than chance). 
This supports the strong form of H2, which states that linguistic dissemination is predictive of growth, even after controlling for the frequency and the other dissemination metrics. 
Subreddit dissemination also shows a mild causal effect on word growth, up to 60\% in the highest treatment quantile.
The other social dissemination metrics prove to have less effect on word growth.

\subsection{Predictive analysis}
\label{sec:results-binary-predict}
We now turn to prediction to determine the utility of linguistic and social dissemination: using the first $k$ months of data, can we predict whether a word will grow or decline in popularity?
This is similar to previous work in predicting the success of lexical innovations~\cite{kooti2012predicting}, but our goal is to compare the relative predictive power of various dissemination metrics, rather than to maximize accuracy.

\begin{figure}
\centering
\includegraphics[width=\columnwidth]{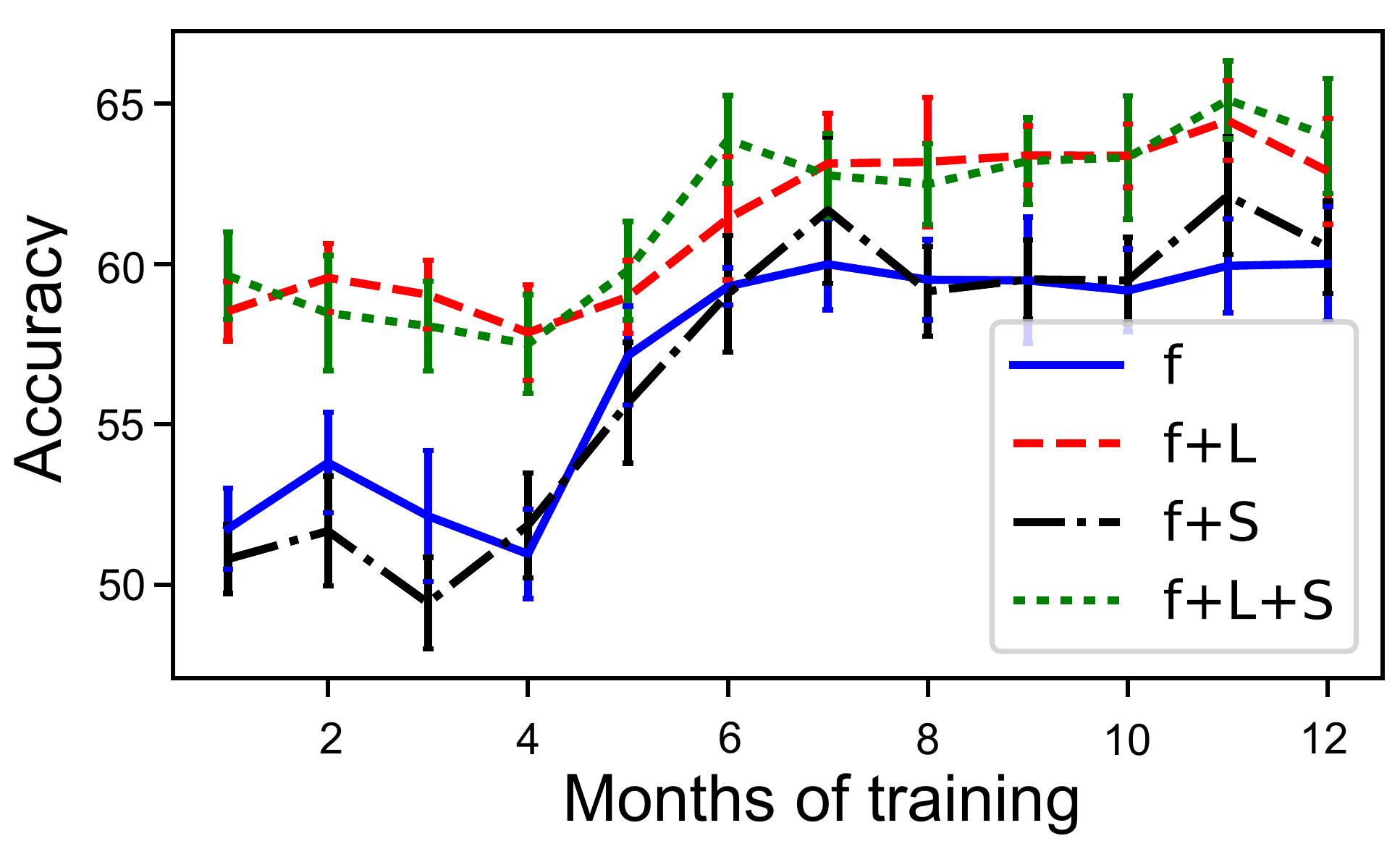}
\caption{Prediction accuracy for different feature sets using $k={1...12}$ months of training data. $f$ indicates frequency-only, $f+L$ frequency plus linguistic dissemination, $f+S$ frequency plus social dissemination, $f+L+S$ all features.}
\label{fig:success_accuracy}
\end{figure}

We use logistic regression with 10-fold cross-validation over four different feature sets: frequency-only (\model{f}), frequency plus linguistic dissemination (\model{f+L}), frequency plus social dissemination (\model{f+S}) and all features (\model{f+L+S}).
Each fold is balanced for classes so that the baseline accuracy is 50\%.
\autoref{fig:success_accuracy} shows that linguistic dissemination provides more predictive power than social dissemination: the accuracy is consistently higher for the models with linguistic dissemination than for the frequency-only and social dissemination models.
The accuracies converge as the training data size increases, which suggests that frequency is a useful predictor if provided sufficient historical trajectory.

\begin{figure}
\centering
\includegraphics[width=\columnwidth]{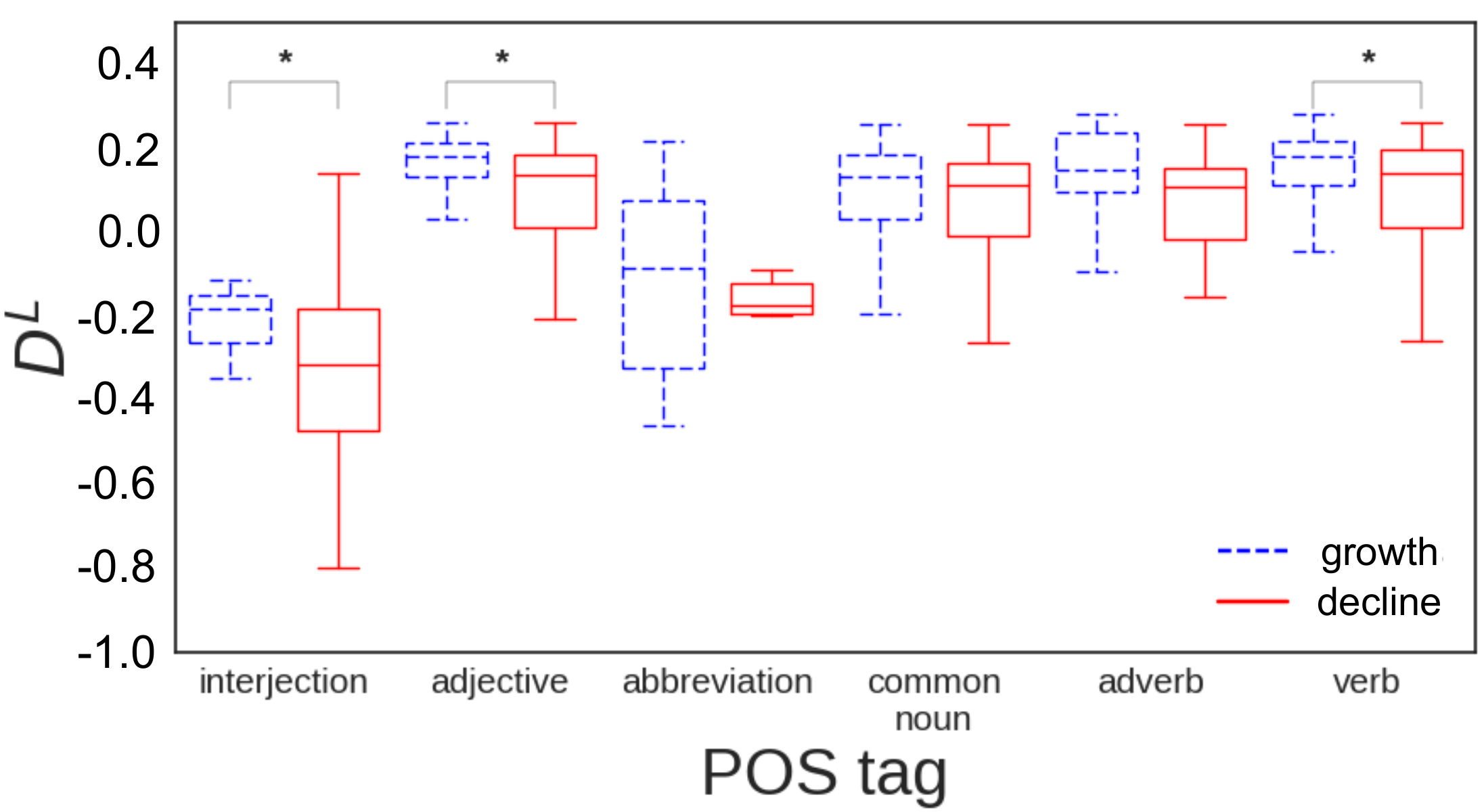}
\caption{Distribution of $D^{L}$ values across growth and decline words, grouped by part of speech tag. 
* indicates $p<0.05$ in one-tailed t-test between growth and decline $D^{L}$ values.}
\label{fig:success_vs_failure_pos_DL_distribution}
\end{figure}

\paragraph{Part-of-speech robustness check}

Considering the uneven distribution of linguistic dissemination across part-of-speech groups (\autoref{fig:pos-cd-dist}), the prediction results may be explained by an imbalance of word categories between the growth and decline words.
This issue is addressed through two robustness checks: within-group comparison and prediction.

First, we compare the distribution of linguistic dissemination values between growth and decline words, grouped by the most common POS tags (computed in \autoref{sec:linguistic_dissemination}).
Each decline word is matched with a growth word based on similar mean frequency in the first $k=12$ months, and their mean linguistic dissemination values during that time period are compared, grouped within POS tag groups.
The differences in~\autoref{fig:success_vs_failure_pos_DL_distribution} show that across all POS tags, the growth words show a tendency toward higher linguistic dissemination with significant ($p<0.05$) differences in the interjections, adjectives and verbs.

Next, we add POS tags as additional features to the frequency-only model in the binary prediction task.
The accuracy of a predictive model with access to frequency and POS features at ${k=1}$ is 54.8\%, which is substantially lower than the accuracy of the model with frequency and linguistic dissemination (cf. \autoref{fig:success_accuracy}).\footnote{Higher $k$ values yield similar results.}
Thus, linguistic dissemination thus contributes predictive power beyond what is contributed by part-of-speech alone.

\subsection{Survival analysis}
\label{sec:results-survival}
Having investigated what separates growth from decline, we now focus on the factors that precede a decline word's ``death'' phase~\cite{drouin2009}.

Predicting the time until a word's decline can be framed as survival analysis~\cite{klein2005}, in which a word is said to ``survive'' until the beginning of its decline phase at split point $\hat{t}$. 
In the Cox proportional hazards model~\cite{david1972}, the hazard of death $\lambda$ at each time $t$ is modeled as a linear function of a vector of predictors,
\begin{equation}
  \lambda_i(t)  = \lambda_0(t) \exp (\mathbf{\beta} \cdot \mathbf{x}_i),
\end{equation}
where $\mathbf{x}_i$ is the vector of predictors for word $i$, and $\mathbf{\beta}$ is the vector of coefficients. Each cell $x_{i,j}$ is set to the mean value of predictor $j$ for word $i$ over the training period $t=\{1 ... k\}$ where $k=3$.

For words which begin to decline in popularity in our dataset, we treat the point of decline as the ``death'' date. The remaining words are viewed as \emph{censored} instances: they may begin to decline in popularity at some point in the future, but this time is outside our frame of observation.
We use frequency, social dissemination and linguistic dissemination as predictors in a Cox regression model.\footnote{Cox regression implemented in the \emph{lifelines} package in Python: \url{https://lifelines.readthedocs.io/en/latest/}.}

\begin{table}[t!]
\small
\centering
\begin{tabular}{l r r r r}
\toprule
  Predictor & $\beta$ & std. error & $Z$ & $p$ \\ \midrule
$\var{f}$ & -0.207 & 0.0492 & -4.21 & *** \\ 
$\var{D^{L}}$ & -0.330 & 0.0385 & -8.56 & *** \\ 
$\var{D^{U}}$ & 0.0053 & 0.0518 & 0.102 & ~ \\ 
$\var{D^{S}}$ & -0.156 & 0.0807 & -1.928 & ~ \\ 
  $\var{D^{T}}$ & 0.0825 & 0.0662 & 1.25 & ~ \\
  \bottomrule
\end{tabular}
\caption{Cox regression results for predicting word death with all predictors (\model{f+L+S}) averaged over first $k=3$ months.
*** indicates $p<0.001$, otherwise $p>0.05$.}
\label{tab:cox_regression_growth_decline_up_to_3}
\end{table}

The estimated coefficients from the regression are shown in \autoref{tab:cox_regression_growth_decline_up_to_3}.
We find a negative coefficient for linguistic dissemination ($\beta=-0.330, p<0.001$), which mirrors the results from \autoref{sec:results-causal}: 
higher $D^{L}$ indicates a lower hazard of word death, and therefore a higher likelihood of survival.
We also find that higher subreddit dissemination has a weak but insignificant correlation with a lower likelihood of word death ($\beta=-0.156, p>0.05$).
Both of these results lend additional support to the strong form of the hypothesis H2.

\begin{figure}[t!]
\centering
\includegraphics[width=0.8\columnwidth]{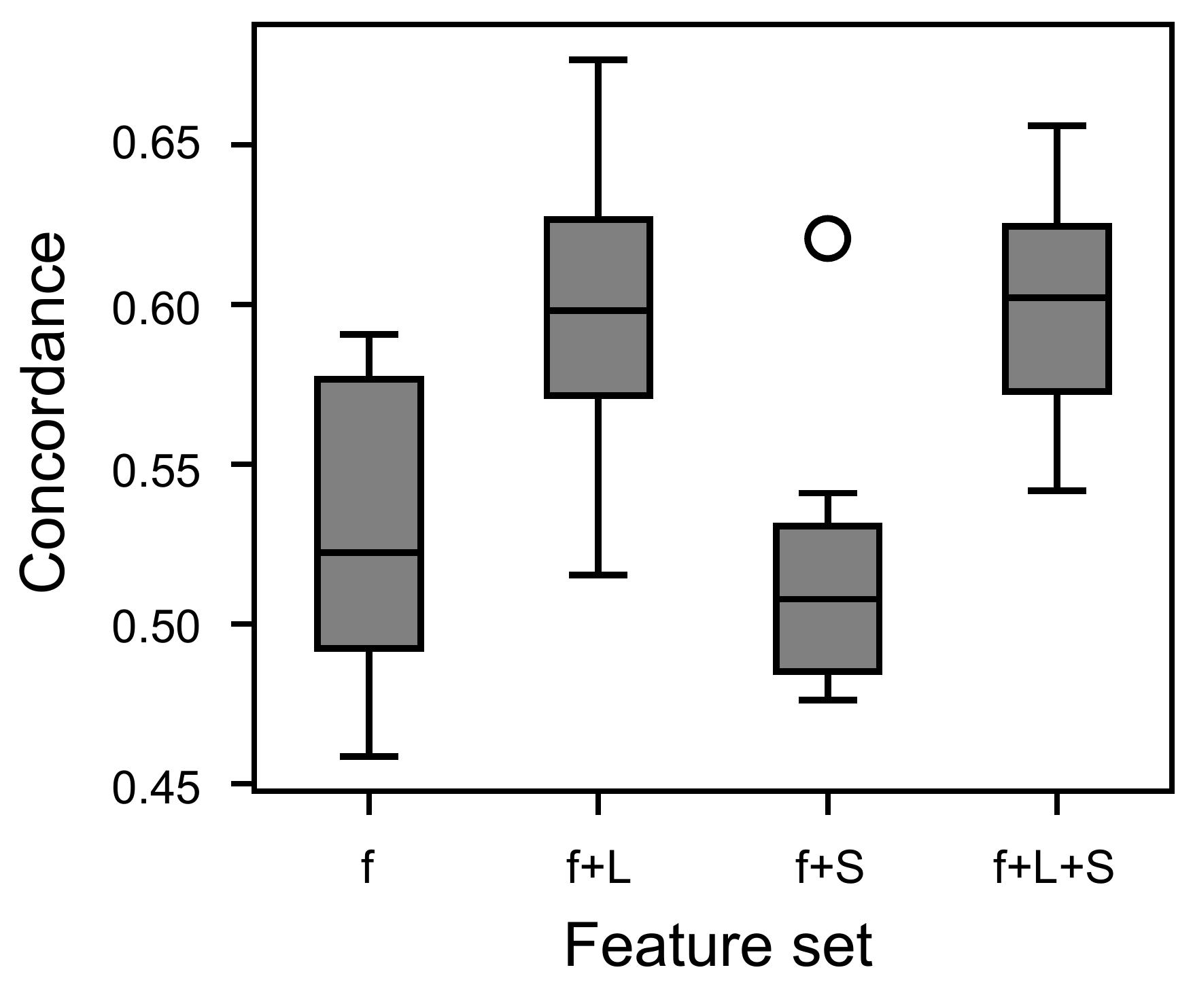}
\caption{Distribution of concordance scores (10-fold cross-validation) of the Cox regression models across feature sets.}
\label{fig:cox_regression_concordance_scores}
\end{figure}

The predictive accuracy of survival analysis can be quantified by a \emph{concordance score}. A score of 1.0 on heldout data indicates that the model perfectly predicts the order of death times; a score of 0.5 indicates that the predictions are no better than a chance ordering. We perform 10-fold cross-validation of the survival analysis model, and plot the results in \autoref{fig:cox_regression_concordance_scores}.
The model with access to linguistic dissemination (\model{f+L}) consistently achieves higher concordance than the baseline frequency-only model (\model{f}), 
($t=4.29, p<0.001$), and the model with all predictors \model{f+L+S} significantly outperforms the model with access only to frequency and social dissemination \model{f+S} ($t=4.64, p<0.001$).
The result is reinforced by testing the goodness-of-fit for each model with model \emph{deviance}, or difference from the null model. 
The \model{f+L} model has lower deviance, i.e. better fit, than the null model ($\chi^{2}=93.3, p < 0.01$), and the \model{f+L+S} does not have a significantly lower deviance than the \model{f+L} model ($\chi^{2}=4.6, p=0.80$), suggesting that adding social dissemination does not significantly improve model fit.

\section{Discussion}



All four analyses support H2: linguistic dissemination was the strongest predictor of monthly frequency changes in growth words, the best differentiator of growth and decline words in causal and predictive tasks, and the most effective warning sign that a word is about to decline.
Linguistic dissemination can be related to theories such as the FUDGE factors~\cite{chesley2010,cook2010neologism,metcalf2004}, in which a word's growth depends on frequency (F), unobtrusiveness (U), diversity of users and situations (D), generation of other forms and meanings (G), and endurance (E). 
Linguistic dissemination provides an example of ``diversity of situation.'' 

The effectiveness of linguistic dissemination is exemplified in pairs of semantically similar growth and decline words.
In the first $k=3$ months of growth, the growth word \example{kinda} has a relatively high ratio of linguistic to frequency ($\frac{D^{L}}{f}=0.270$) as compared with the semantically similar decline word \example{sorta} (0.055).
This pattern holds for other pairs of semantically similar growth/decline words: \example{fuckwit} and \example{fuckboy}; \example{lolno} and \example{lmao}; \example{yup} and \example{yas}.
While not exhaustive, such a trend suggests that the growth words were able to reach a wider range of lexical contexts and therefore succeed where the decline words failed.

Regarding H1, we generally found a positive role for social dissemination as well, although these results were not consistent across all metrics and tests, particularly in the survival analysis.
This matches the conclusion from~\newcite{garley2012}, who argued that social dissemination is less predictive of word adoption than~\newcite{altmann2011} originally suggested.
One possible explanation is the inclusion of word categories such as proper nouns in the analysis of \newcite{altmann2011}; the adoption of such terms may rely on social dynamics more than the adoption of nonstandard terms.
The lower predictive power of thread and user dissemination is also interesting and suggests that subreddits are more socially salient in terms of exposing nonstandard words to potential adopters.

\paragraph{Limitations}

One limitation in the study was the exclusion of orthographic and morphological features such as affixation, which has been noted as a predictor of word growth~\cite{kershaw2016}.
Future work should incorporate these features as additional predictors.
Our study also omitted borrowings, unlike prior work in word adoption that has focused on borrowings~\cite{chesley2010,garley2012}.
Our early language-filtering steps eliminated most non-English words from the vocabulary, although it would have been interesting to examine loanword use in English-language posts.
Finally, our study was limited by the focus on nonstandard words rather than memetic phrases (e.g., \example{like a boss}) which may show a similar correlation between dissemination, growth and decline~\cite{bybee2006}.

\paragraph{Future work} 
We approximate linguistic dissemination using trigram counts, because they are easy to compute and they generalize across word categories. 
In future work, a more sophisticated approach might estimate linguistic dissemination with syntactic features such as appearance across different phrase heads~\cite{kroch1989,ito2003} or across nouns of different semantic classes~\cite{darcy2015}.
Future work should also investigate more semantically-aware definitions of linguistic dissemination.
The existence of semantic ``neighbors'' occurring in similar contexts (e.g., the influence of standard intensifier \example{very} on nonstandard intensifier \example{af}) may prevent a new word from reaching widespread popularity~\cite{grieve2018}.



\section*{Acknowledgments}
The authors thank the anonymous reviewers, the audience at NWAV 46 for discussion of an early presentation of this study, and the members of Georgia Tech's Computational Linguistics Lab for their help throughout the project. This research was supported by NSF award IIS-1452443, AFOSR award FA9550-14-1-0379, and NIH award R01-GM112697-03.

\bibliography{references}

\begin{thebibliography}{44}
\expandafter\ifx\csname natexlab\endcsname\relax\def\natexlab#1{#1}\fi

\bibitem[{Altmann et~al.(2011)Altmann, Pierrehumbert, and Motter}]{altmann2011}
Eduardo Altmann, Janet Pierrehumbert, and Adilson Motter. 2011.
\newblock {Niche as a determinant of word fate in online groups}.
\newblock \emph{PLoS ONE}, 6(5):1--12.

\bibitem[{Androutsopoulos(2011)}]{androutsopoulos2011}
Jannis Androutsopoulos. 2011.
\newblock {Language change and digital media: a review of conceptions and
  evidence}.
\newblock In Kristiansen Tore and Nikolas Coupland, editors, \emph{Standard
  Languages and Language Standards in a Changing Europe}, pages 145--159. Novus
  Press, Oslo.

\bibitem[{Angrist et~al.(1996)Angrist, Imbens, and Rubin}]{angrist1996}
Joshua~D Angrist, Guido~W Imbens, and Donald~B Rubin. 1996.
\newblock {Identification of Causal Effects Using Instrumental Variables}.
\newblock \emph{Source Journal of the American Statistical Association},
  91(434):444--455.

\bibitem[{Bakshy et~al.(2011)Bakshy, Hofman, Mason, and
  Watts}]{bakshy2011everyone}
Eytan Bakshy, Jake Hofman, Winter Mason, and Duncan Watts. 2011.
\newblock {Everyone's an influencer: quantifying influence on Twitter}.
\newblock In \emph{Proceedings of the International Conference on Web Search
  and Data Mining}, pages 65--74.

\bibitem[{Bucholtz(1999)}]{bucholtz1999}
Mary Bucholtz. 1999.
\newblock {"Why be normal?": Language and identity practices in a community of
  nerd girls}.
\newblock \emph{Language in Society}, 28:203--223.

\bibitem[{Bybee(2006)}]{bybee2006}
Joan~L. Bybee. 2006.
\newblock {From Usage to Grammar: The Mind's Response to Repetition}.
\newblock \emph{Language}, 82(4):711--733.

\bibitem[{Cacoullos and Walker(2009)}]{cacoullos2009}
Rena~Torres Cacoullos and James Walker. 2009.
\newblock {The Present of the English Future: Grammatical Variation and
  Collocations in Discourse}.
\newblock \emph{Language}, 85(2):321--354.

\bibitem[{Chesley and Baayen(2010)}]{chesley2010}
Paula Chesley and Harald Baayen. 2010.
\newblock {Predicting new words from newer words: Lexical borrowings in
  French}.
\newblock \emph{Linguistics}, 48(6):1343--1374.

\bibitem[{Cook(2010)}]{cook2010neologism}
Paul Cook. 2010.
\newblock \emph{Exploiting linguistic knowledge to infer properties of
  neologisms}.
\newblock Ph.D. thesis, University of Toronto.

\bibitem[{Cook and Stevenson(2010)}]{cook2010blend}
Paul Cook and Suzanne Stevenson. 2010.
\newblock {Automatically identifying the source words of lexical blends in
  English}.
\newblock \emph{Computational Linguistics}, 36(1):129--149.

\bibitem[{Cox(1972)}]{david1972}
David Cox. 1972.
\newblock Regression models and life tables.
\newblock \emph{Journal of the Royal Statistical Society}, 34:187--220.

\bibitem[{Danescu-Niculescu-Mizil et~al.(2013)Danescu-Niculescu-Mizil, West,
  Jurafsky, and Potts}]{danescu2013}
Cristian Danescu-Niculescu-Mizil, Robert West, Dan Jurafsky, and Christopher
  Potts. 2013.
\newblock {No Country for Old Members: User Lifecycle and Linguistic Change in
  Online Communities}.
\newblock \emph{Proceedings of the International Conference on World Wide Web},
  pages 307--317.

\bibitem[{D'Arcy and Tagliamonte(2015)}]{darcy2015}
Alexandra D'Arcy and Sali Tagliamonte. 2015.
\newblock {Not always variable: probing the vernacular grammar}.
\newblock \emph{Language Variation and Change}, 27(3):255--285.

\bibitem[{Drouin and Dury(2009)}]{drouin2009}
Patrick Drouin and Pascaline Dury. 2009.
\newblock When terms disappear from a specialized lexicon: A semi-automatic
  investigation into necrology.
\newblock In \emph{{Actes de la conf{\'e}rence internationale ``Language for
  Special Purposes''}}.

\bibitem[{Dumas and Lighter(1978)}]{dumas1978}
BK~Dumas and Jonathan Lighter. 1978.
\newblock {Is slang a word for linguists?}
\newblock \emph{American Speech}, 53(1):5--17.

\bibitem[{Egghe(2007)}]{egghe2007}
Leo Egghe. 2007.
\newblock {Untangling Herdan's law and Heaps' Law: Mathematical and informetric
  arguments}.
\newblock \emph{Journal of the American Society for Information Science and
  Technology}, 58(5):702--709.

\bibitem[{Gaffney and Matias(2018)}]{gaffney2018}
Devin Gaffney and J.~Nathan Matias. 2018.
\newblock {Caveat emptor, computational social science: Large-scale missing
  data in a widely-published reddit corpus}.
\newblock \emph{PLoS ONE}, 13(7).

\bibitem[{Garley and Hockenmaier(2012)}]{garley2012}
Matt Garley and Julia Hockenmaier. 2012.
\newblock {Beefmoves: dissemination, diversity, and dynamics of English
  borrowings in a German hip hop forum}.
\newblock In \emph{Proceedings of the Association of Computational
  Linguistics}, pages 135--139.

\bibitem[{Gilbert(2013)}]{gilbert2013}
Eric Gilbert. 2013.
\newblock {Widespread Underprovision on Reddit}.
\newblock In \emph{Proceedings of the Conference on Computer-Supported
  Cooperative Work}, pages 803--808.

\bibitem[{Gimpel et~al.(2011)Gimpel, Schneider, O'Connor, Das, Mills,
  Eisenstein, Heilman, Yogatama, Flanigan, and Smith}]{gimpel2011}
Kevin Gimpel, Nathan Schneider, Brendan O'Connor, Dipanjan Das, Daniel Mills,
  Jacob Eisenstein, Michael Heilman, Dani Yogatama, Jeffrey Flanigan, and Noah
  Smith. 2011.
\newblock {Part-of-speech tagging for Twitter: Annotation, features, and
  experiments}.
\newblock In \emph{Proceedings of the 49th Annual Meeting of the Association
  for Computational Linguistics}, pages 42--47.

\bibitem[{Grieve(2018)}]{grieve2018}
Jack Grieve. 2018.
\newblock {Natural selection in the modern English Lexicon}.
\newblock In \emph{International Conference on Language Evolution}, pages
  153--157.

\bibitem[{Grieve et~al.(2016)Grieve, Nini, and Guo}]{grieve2016}
Jack Grieve, Andrea Nini, and Diansheng Guo. 2016.
\newblock {Analyzing lexical emergence in Modern American English online}.
\newblock \emph{English Language and Linguistics}, 20(2):1--29.

\bibitem[{Hessel et~al.(2016)Hessel, Tan, and Lee}]{hessel2016}
Jack Hessel, Chenhao Tan, and Lillian Lee. 2016.
\newblock {Science, AskScience, and BadScience: On the coexistence of highly
  related communities}.
\newblock In \emph{Proceedings of the International Conference on Social and
  Web Media}, pages 171--180.

\bibitem[{Hirano and Imbens(2004)}]{hirano2004}
Keisuke Hirano and Guido~W Imbens. 2004.
\newblock The propensity score with continuous treatments.
\newblock In Andrew Gelman and Xiao-Li Meng, editors, \emph{{Applied Bayesian
  modeling and causal inference from incomplete-data perspectives}}, pages
  73--84. Wiley, Chichester.

\bibitem[{Imbens(2000)}]{imbens2000}
Guido~W Imbens. 2000.
\newblock The role of the propensity score in estimating dose-response
  functions.
\newblock \emph{Biometrika}, 87(3):706--710.

\bibitem[{Ito and Tagliamonte(2003)}]{ito2003}
Rika Ito and Sali Tagliamonte. 2003.
\newblock {Well weird, right dodgy, very strange, really cool: Layering and
  recycling in English intensifiers}.
\newblock \emph{Language in Society}, 32:257--279.

\bibitem[{Kershaw et~al.(2016)Kershaw, Rowe, and Stacey}]{kershaw2016}
Daniel Kershaw, Matthew Rowe, and Patrick Stacey. 2016.
\newblock {Towards Modelling Language Innovation Acceptance in Online Social
  Networks}.
\newblock \emph{Proceedings of the ACM International Conference on Web Search
  and Data Mining}, pages 553--562.

\bibitem[{Klein and Moeschberger(2005)}]{klein2005}
John Klein and Melvin Moeschberger. 2005.
\newblock \emph{Survival analysis: techniques for censored and truncated data}.
\newblock Springer Science \& Business Media, New York.

\bibitem[{Kooti et~al.(2012)Kooti, Mason, Gummadi, and
  Cha}]{kooti2012predicting}
Farshad Kooti, Winter~A Mason, Krishna~P Gummadi, and Meeyoung Cha. 2012.
\newblock {Predicting emerging social conventions in online social networks}.
\newblock In \emph{Proceedings of the International Conference on Information
  and Knowledge Management}, pages 445--454.

\bibitem[{Kroch(1989)}]{kroch1989}
Anthony~S Kroch. 1989.
\newblock Reflexes of grammar in patterns of language change.
\newblock \emph{Language Variation and Change}, 1(3):199--244.

\bibitem[{Kruskal(1987)}]{kruskal1987}
William Kruskal. 1987.
\newblock Relative importance by averaging over orderings.
\newblock \emph{The American Statistician}, 41(1):6--10.

\bibitem[{Kulkarni and Wang(2018)}]{kulkarni2018}
Vivek Kulkarni and William~Yang Wang. 2018.
\newblock {Simple Models for Word Formation in English Slang}.
\newblock In \emph{In Proceedings of the North American Association of
  Computational Linguistics}, pages 1424--1434.

\bibitem[{Labov(2007)}]{labov2007}
William Labov. 2007.
\newblock {Transmission and Diffusion}.
\newblock \emph{Language}, 83(2):344--387.

\bibitem[{Lui and Baldwin(2012)}]{lui2012}
Marco Lui and Timothy Baldwin. 2012.
\newblock {langid. py: An off-the-shelf language identification tool}.
\newblock In \emph{Proceedings of the Association of Computational
  Linguistics}, pages 25--30.

\bibitem[{MacWhinney(1989)}]{macwhinney1989}
Brian MacWhinney. 1989.
\newblock {Competition and Lexical Categorization}.
\newblock \emph{Linguistic Categorization}, pages 195--242.

\bibitem[{Metcalf(2004)}]{metcalf2004}
Allan Metcalf. 2004.
\newblock \emph{{Predicting new words: The secrets of their success}}.
\newblock Houghton Mifflin, New York.

\bibitem[{Partington(1993)}]{partington1993}
Alan Partington. 1993.
\newblock {Corpus evidence of language change: The case of the intensifier}.
\newblock In Mona Baker, Gill Francis, and Elena Tognini-Bonelli, editors,
  \emph{Text and Technology: In Honour of John Sinclair}, pages 177--192. John
  Benjamins Publishing, Philadelphia.

\bibitem[{Romero et~al.(2011)Romero, Meeder, and
  Kleinberg}]{romero2011differences}
Daniel Romero, Brendan Meeder, and Jon Kleinberg. 2011.
\newblock {Differences in the mechanics of information diffusion across topics:
  idioms, political hashtags, and complex contagion on Twitter}.
\newblock In \emph{Proceedings of the International Conference on World Wide
  Web}, pages 695--704. ACM.

\bibitem[{Squires(2010)}]{squires2010enregistering}
Lauren Squires. 2010.
\newblock Enregistering internet language.
\newblock \emph{Language in Society}, 39:457--492.

\bibitem[{Tagliamonte and Denis(2008)}]{tagliamonte2008}
Sali Tagliamonte and Derek Denis. 2008.
\newblock {Linguistic ruin? LOL! Instant messaging and teen language}.
\newblock \emph{American Speech}, 83(1):3--34.

\bibitem[{Tan and Lee(2015)}]{tan2015}
Chenhao Tan and Lillian Lee. 2015.
\newblock All who wander: On the prevalence and characteristics of
  multi-community engagement.
\newblock In \emph{Proceedings of the International Conference on World Wide
  Web}, pages 1056--1066.

\bibitem[{Tonidandel et~al.(2009)Tonidandel, LeBreton, and
  Johnson}]{tonidandel2009}
Scott Tonidandel, James LeBreton, and Jeff Johnson. 2009.
\newblock Determining the statistical significance of relative weights.
\newblock \emph{Psychological methods}, 14(4):387.

\bibitem[{Tredici and Fern{\'{a}}ndez(2018)}]{tredici2018}
Marco~Del Tredici and Raquel Fern{\'{a}}ndez. 2018.
\newblock {The Road to Success: Assessing the Fate of Linguistic Innovations in
  Online Communities}.
\newblock In \emph{Proceedings of the International Conference on Computational
  Linguistics}, pages 1591--1603.

\bibitem[{Tsur and Rappoport(2015)}]{tsur2015}
Oren Tsur and Ari Rappoport. 2015.
\newblock {Don't Let Me Be {\#}Misunderstood: Linguistically Motivated
  Algorithm for Predicting the Popularity of Textual Memes}.
\newblock In \emph{Proceedings of the International Conference on Social and
  Web Media}, pages 426--435.

\end{thebibliography}
\bibliographystyle{acl_natbib_nourl}

\end{document}